\documentclass[conference]{IEEEtran}
\IEEEoverridecommandlockouts
\usepackage{cite}
\usepackage{amsmath,amssymb,amsfonts}
\usepackage{algorithmic}
\usepackage{graphicx}
\usepackage{hyperref}
\usepackage{textcomp}
\usepackage{tikz}
\usepackage{multirow}
\usepackage{xcolor}
\usepackage{subfig}
\def\BibTeX{{\rm B\kern-.05em{\sc i\kern-.025em b}\kern-.08em
    T\kern-.1667em\lower.7ex\hbox{E}\kern-.125emX}}

\begin{document}

\title{A Study of the Plausibility of Attention between RNN Encoders in Natural Language Inference}

\author{\IEEEauthorblockN{Duc Hau Nguyen}
\IEEEauthorblockA{\textit{IRISA, CNRS, INSA Rennes} \\
Rennes, France \\
duc-hau.nguyen@irisa.fr}
\and
\IEEEauthorblockN{Guillaume Gravier}
\IEEEauthorblockA{\textit{IRISA, CNRS} \\
Rennes, France \\
guillaume.gravier@irisa.fr}
\and
\IEEEauthorblockN{Pascale S\'{e}billot}
\IEEEauthorblockA{\textit{IRISA, INSA Rennes} \\
Rennes, France \\
pascale.sebillot@irisa.fr}
\thanks{Work partially funded by grant ANR-19-CE38-0011-03 from the French national research agency (ANR).}
}

\maketitle

\begin{abstract}
Attention maps in neural models for NLP are appealing to explain the
decision made by a model, hopefully emphasizing words that justify the
decision. While many empirical studies hint that attention maps can
provide such justification from the analysis of sound examples, only a
few assess the plausibility of explanations based on attention maps,
i.e., the usefulness of attention maps for humans to understand the
decision. These studies furthermore focus on text classification. In
this paper, we report on a preliminary assessment of attention maps in
a sentence comparison task, namely natural language inference. We
compare the cross-attention weights between two RNN encoders with
human-based and heuristic-based annotations on the eSNLI corpus. We
show that the heuristic reasonably correlates with human annotations
and can thus facilitate evaluation of plausible explanations in
sentence comparison tasks. Raw attention weights however remain only
loosely related to a plausible explanation.

\end{abstract}


\section{Introduction}
\label{sec:introduction}

Attention mechanisms in deep learning have gained huge popularity in the past 5 years due to the benefits on performance across many tasks. In natural language processing (NLP) in particular, these mechanisms are now widely used, either as an attention layer on top of a recurrent network~\cite{luong_effective_2015, bahdanau_neural_2016, chen_recurrent_2017} or directly as a self-attention model~\cite{vaswani_attention_2017, devlin_bert_2019, shen_disan_2018 }.

In addition to improved performance, attention weights have also attracted interest as a way to help human understand how deep learning models process data and make decisions~\cite{wang_learning_2015,  wang_attention-based_2016, ghaeini_interpreting_2018}. Typically, in NLP, an attention mechanism provides a weight on each input word, presumably proportional to the word's relevance regarding the decision, and weights can be visualized as a heat map, known as an {\em attention map}.

However, qualifying the usefulness for a human to understand the decision that was made by a model---a concept globally known as {\em explanability}---remains a challenge. The challenge arises from two concerns: on the one hand, explanability isn't directly related to performance on the end task; on the other hand, the cost of post-hoc user studies or the lack of annotated data for attention layers impose severe limits, as attention maps are usually one of several layers within a complex architecture. Moreover, regardless of performance, most of the evaluations of attention maps for explanability in NLP were done empirically at the sentence level, with models that were primarily designed for performance rather than for explanability.

It is now well-known that a model with a good performance on a certain task does not guarantee a good explanation, which makes the end-task performance a poor proxy to assess explanability. Recent works, such as~\cite{jain_attention_2019, wiegreffe_attention_2019, serrano_is_2019}, disproved the uniqueness of explanation for the same prediction by showing that there exists an adversarial model maintaining similar prediction but having a completely different attention map. 

Evaluation of the usefulness of attention maps regarding explanability must therefore focus on human assessment of the attention map itself. Two natural approaches can be considered to this end: posterior user studies of attention maps, which are costly and hardly provide sound criteria to train attention mechanisms with explanability in mind; prior annotation of important input token/words that should be emphasized to explain the decision.

In both cases, it is nevertheless unclear whether human judgement necessarily provides a good baseline for explanation, depending on what is meant by explanation. Interestingly, the notions of faithfulness and plausibility were recently introduced~\cite{jacovi_towards_2020}. Briefly speaking, faithfulness reflects how much an attention map "reflects the model's reasoning process"---in other words, how a model makes the decision with the same chosen words or a chosen part of input in general---, while plausibility reflects how useful the attention map is to humans.

Most recent studies on explanability derived from attention maps assess the faithfulness of the attention mechanism, e.g., qualitatively looking at attention weights on a few examples \cite{xu_show_2015, choi_retain_2016, ghaeini_interpreting_2018, lin_structured_2017}.
Addressing plausibility is however much less studied, because of the lack of ground truth and the difficulty to define one. In one of the few studies on plausibility of attention maps in NLP~\cite{mullenbach_explainable_2018}, annotators were asked to classify attention maps as informative or not in a medical text classification task.

In this paper, we investigate the factors that influence the plausibility of an attention mechanism between two LSTM recurrent encoders on a natural language inference (NLI) task where two sentences are compared to detect entailment or contradiction. The choice of the task responds to two criteria apart from novelty: it is significantly more complex than text classification, as considered in~\cite{mullenbach_explainable_2018}; it is close to an information retrieval task where links between text fragments must be established. This last consideration is important to us as learning to link two sentences is critical in many applications, often requires plausible explanations for users to understand why a link is proposed, and hardly comes with a ground truth explanation from which an explanation model can be learned.

Taking NLI as an emblematic task where two sentences have to be compared, we take advantage of human annotations within the eSNLI dataset~\cite{camburu_e-snli_2018}, an extension of the standard Stanford Natural Language Inference dataset~\cite{bowman_large_2015} that comes with human annotations of the words that should be highlighted in the premise and hypothesis sentences. We compare attention weights between two RNN encoders with human annotations and with a simple heuristic, close to the one proposed in~\cite{mullenbach_explainable_2018} for text classification. Although based on a different criterion than the annotation instructions in eSNLI, the heuristic is fully justified for the entailment task, in particular to get plausible explanations in the case of entailment. We also experimentally show that the heuristic reasonably correlates with human annotations and can thus facilitate evaluation of plausible explanations in sentence comparison tasks.


\section{Related Works}
\label{sec:related}

Originally, attention mechanisms were introduced in machine
translation to help the focus on contextual words in sequence
encoding~\cite{bahdanau_neural_2016, luong_effective_2015}. A brief
visualization from the original papers demonstrating responsible
tokens for each translation words has inspired other studies to use
attention maps~\cite{ghaeini_interpreting_2018, lin_structured_2017}
to demonstrate influential parts of each instance's decision (also
referred as local explanation \cite{ghaeini_interpreting_2018,
  gilpin_explaining_2018}). Similar approaches were taken for
self-attention models.

Following this trend, \cite{jain_attention_2019} questioned the
validity of the attention map and resulted in disproving
the faithfulness in explanation in some tasks. In other
words, we can have a different attention map to get the same
prediction result. \cite{serrano_is_2019} has disproved that an
attention map does not have the linear contribution to the final
prediction, as it can mask out important words and still get the
approximate prediction. To explain this phenomenon,
\cite{wiegreffe_attention_2019} has introduced the concept of faithfulness when trying to evaluate an attention mechanism, which
refers to the capability that an attention map can reflect the model
reasoning.  The authors argued that the attention mechanism does
reflect the model's deduction process but is not necessarily
comprehensible to human. Another factor is the contribution of the
attention layer in the task: if this layer is replaceable, then it is
unlikely to have any explanation
property. \cite{vashishth_attention_2019} has bolstered this argument,
demonstrating that in single sequence inference tasks (sentiment or
topic classifications for example), the attention score becomes just a
gate unit used in a convolution model~\cite{dauphin_language_2017}.

While the faithfulness of the model has been the subject of several works, the plausibility of attention maps is still rarely studied to our knowledge. While there is a direct way to measure plausibility of attention map in computer vision-related tasks, as in \cite{wu_faithful_2019} for example, most of NLP corpus does not have a direct way to evaluate against human-annotated ground truth. Instead,  \cite{nguyen_comparing_2018} studies the usefulness of attention map in a text classification task, asking workers to rank how much he or she is confident to find the correct label given the attention map. 
Others, such as \cite{mullenbach_explainable_2018}, tried to evaluate the plausibility of their heuristic map, asking physicians to rank it against three other heuristics. The study here again relies on text classification tasks. Despite the claim of a satisfactory explanability of their model, result showed that a heuristic map based on similarity between words is more informative than the attention map is.

In this paper, we therefore extend the study of the plausibility of attention maps to a sentence comparison task---natural language inference, not restricting ourselves to a specific domain and taking advantage of existing human annotations in the eSNLI dataset.

\section{Reference Annotations in NLI}
\label{sec:nli}

Natural language inference, also known as recognizing textual
entailment, considers two sentences, namely a premise and an hypothesis, and consists in determining whether the
hypothesis is entailed by or contradicts the premise, or if there is
neither entailment nor contradiction. This typically results in a three
class classification problem.

In this work, we employ the eSNLI corpus, an extension of the standard
SNLI dataset where relevant words for a plausible explanation of the
relation between the two sentences are highlighted by
annotators. Annotators were asked to ``focus on the non-obvious
elements that induce the given relation, and not the parts of the
premise that are repeated identically in the
hypothesis''---see~\cite{camburu_e-snli_2018}, Sec.\ 3 for
details. Presumably, the words highlighted by the annotators, denoted
as highlight map, are thus the ones we would like a plausible
attention-based explanation model to emphasize. Highlight maps are
provided for the entailment and contradiction classes. We focus mostly
on the former in this study, sticking to the information retrieval
scenario described in the introduction.

The eSNLI dataset provides approx.\ 180,000 pairs of sentences for each
class in the train set, and approx.\ 3,000 pairs per class in the dev
and test sets respectively. In these two last sets, the rate of
out-of-vocabulary words (i.e., words not appearing in the train set)
is around 5.5\,\% (resp.\ 6\,\%) without (resp.\ with)
lemmatization. We report in \autoref{table:pos-description} a brief
analysis of the highlighted words in terms of their part-of-speech
(POS) category. Note that a large part of highlighted words are among verb, noun and adjective, which suggests that certain grammatical categories offer better plausibility. Overall, 18\,\% of the words were highlighted by the annotators. 




\begin{table}[tb!]
\caption{Statistics on the POS category of highlighted words.}
\label{table:pos-description}
\begin{center}
\begin{tabular}{|c|c|c|c|c|}
\hline
\multicolumn{2}{|c|}{} & \textbf{\textit{train}}& \textbf{\textit{dev}}& \textbf{\textit{test}} \\
\hline

\multirow{7}{*}{\%~POS tag} & VERB & 21.90\% & 21.20\% & 20.8\%  \\
& NOUN & 49.18\% & 48.4\% & 43.92\%  \\
& ADJ & 9.08\% & 9.14\% & 7.98\%\\
& NUM & 2.89\% & 1.94\% & 1.36\% \\
& ADP & 0.15\% & 2.31\% & 7.55\% \\
& DET & 15,52\% & 9.34\% &  8.30\% \\

\cline{2-5}
&\%(VERB+NOUN+ADJ) & \textbf{73.42\%} & \textbf{78.75\%} & \textbf{72.70\%} \\
\hline

\multicolumn{2}{|c|}{\%~of highlight words} & 18.00\% & 18.21\% & 18,.25\% \\

\hline

\end{tabular}
\end{center}
\end{table}

Finally, let us note that the SNLI corpus is known to have artifacts,
where some lexical fields appear mostly in one class. As reported
in~\cite{gururangan_annotation_2018}, words related to animals and
outdoor are much more frequent in entailment than in the other two
classes and a simple model extracting terms in the hypothesis can
predict well the class. Far from being an issue, this bias is
beneficial to us as it enables to distinguish between faithfulness
(how much the attention map explains the model's reasoning) and
plausibility (how much the attention map is useful for a human to
interpret the decision). Clearly, if attention maps focus on the
model's reasoning in this bias context, they should be poorly
plausible unless one's aware of the bias in the data. 

\section{A Heuristic-based Plausible Attention Map}
\label{sec:heuristic}

Apart from human annotations in the eSNLI corpus, we designed a
heuristic-based approach to plausible attention maps inspired
from~\cite{rocktaschel_reasoning_2015}. The rationale for using a
heuristic that can be computed from the input sentences is that if
valid, i.e., correlated with human annotations, it might serve as a
criterion to drive the model's attention weights towards a plausible
explanation. As of now, the heuristic is used for evaluation purposes.

The heuristic that we propose focuses on the entailment class where the
premise and hypothesis sentences are strongly related (i.e., we can assert
that hypothesis is true given the premise) and seeks to highlight words
that are close in meaning between the two sentences. Note that the
same heuristic might be justified for the contradiction class, however
not for the neutral one. In the case of entailment, it seems a
reasonable assumption to state that plausibility of the explanation
is related to words with close meanings between the two sentences, in
particular verbs, nouns and adjectives.

Formally, let $w_i, \;\; i \in [1, n]$ denote a word token in one text
sequence of length $n$ and $v_j, \;\; j \in [1, m]$ a word token in
the other text sequence of length $m$.  The intuition is that we
measure how relevant each word $w_i$ is by comparing it to every word
$v_j$ in the other sentence. We note a function $\mbox{\it
  similarity}$ that scores the similarity between two words and sum
over the similarity with respect to $w_i$ to get its relative strength
among other words in the sequence. A "good" word should have a better
score, since it has short distant to many words in the other side.

A heuristic score $h(w_i)$ is defined by applying a normalization to
reduce the magnitude of $w_i$ in long sentences. Since words may have
more relatives in longer sentence, the sum tends to be
higher. Denoting as $\sigma$ the normalizing function, the heuristic
is given as
\begin{equation}
h(w_i)=\sigma(\sum_{v_j \notin S}\mbox{\it similarity}(w_i,
v_j)) \label{eq:heuristic} \enspace ,
\end{equation}
where $S$ is the set of stop words. We used the cosine similarity
between embeddings of the words as similarity function. 
The normalizing function $\sigma$ is a rescaling function in the range $[0, 1]$, i.e.,
\begin{equation}
\sigma(x) = \frac{x - \min(x)}{\max(x) - \min(x)} \label{eq:normalizing} \enspace .
\end{equation}

\begin{figure}[b!]
\centerline{\includegraphics[width=0.9\columnwidth]{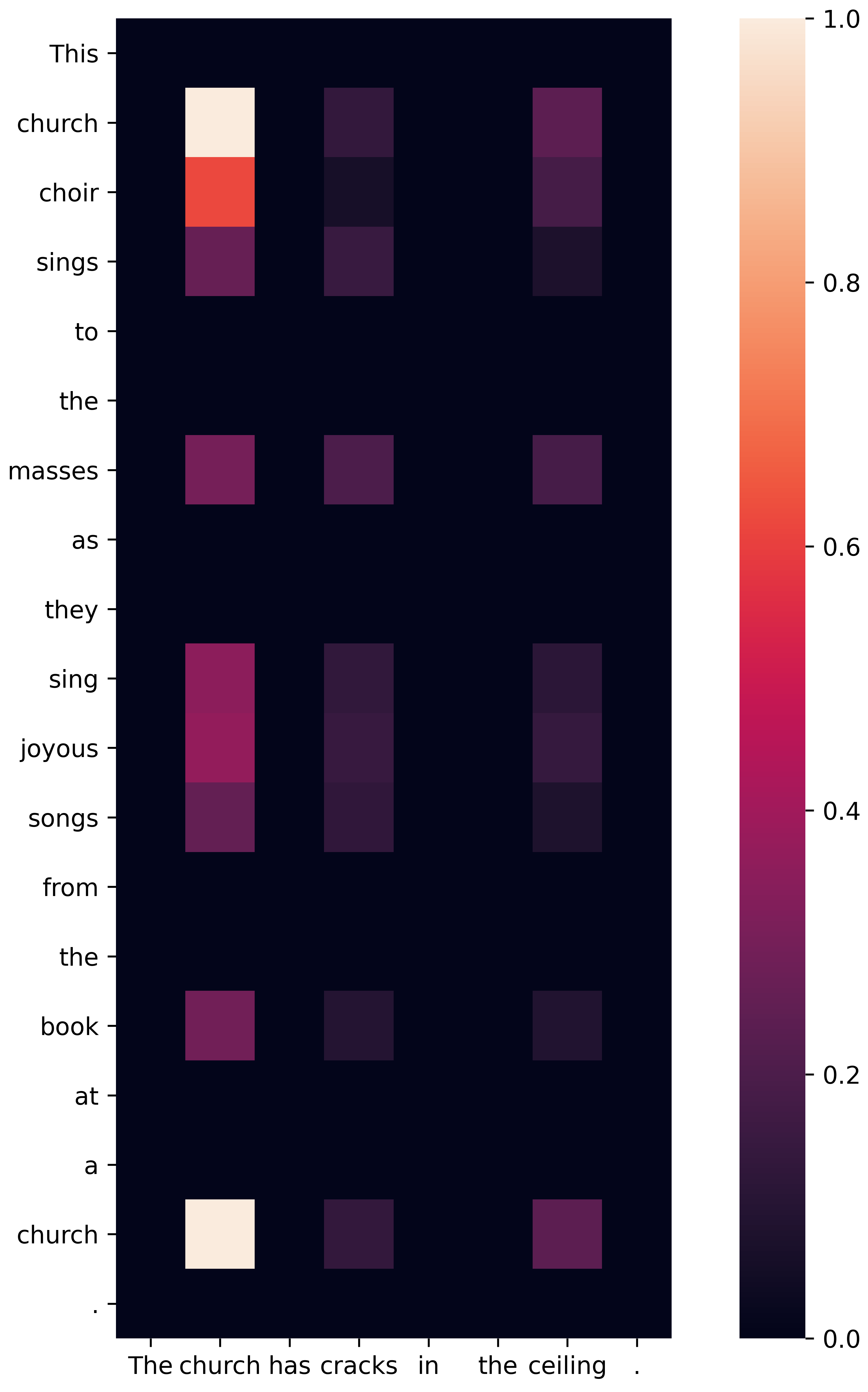}}
\caption{Example of a heuristic-based plausible attention map.}
\label{fig:demo-heuristic}
\end{figure}

A remark is that the heuristic is similar to the "word by word"
attention \cite{rocktaschel_reasoning_2015}, but we rather use the
similarity dot function as
in~\cite{luong_effective_2015}. Fig.~\ref{fig:demo-heuristic}
illustrates the heuristic attention map on an example of a sentence
pair from the corpus: each entry in the attention map is set to
$\mbox{\it similarity}(w_i, v_j) \in [0, 1]$, a high value indicating
a relevant word. Stop words are masked out.

\section{Model-based Attention Map}
\label{model}

\begin{figure}
\centerline{\includegraphics[width=0.8\columnwidth]{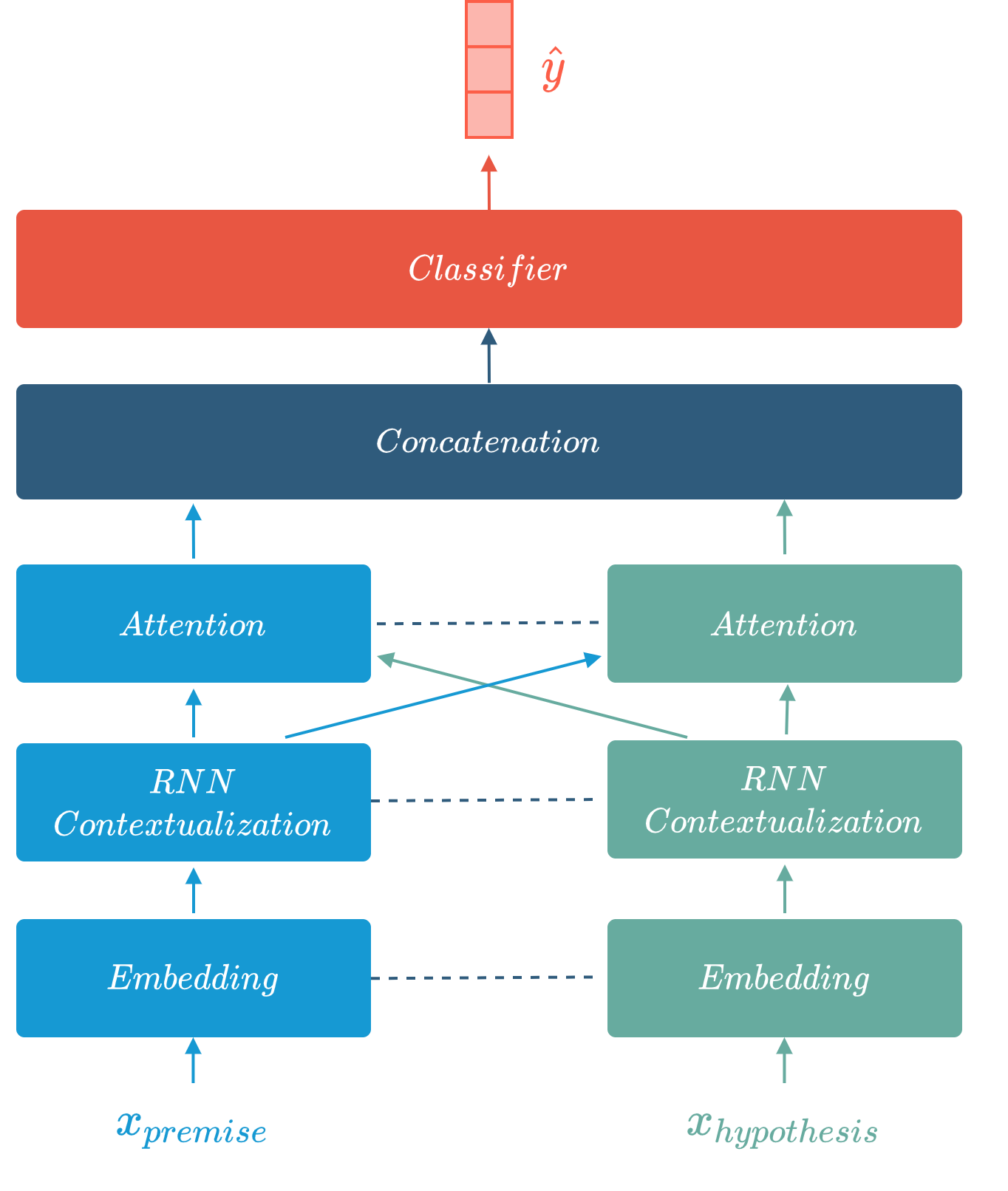}}
\caption{Overall model architecture. The dash lines indicate that layers share the weights; in other words, we use the same first 3 layers ($Embedding$,  $Contextualization$, $Attention$) for both $premise$ and $hypothesis$.}
\label{fig:model}
\end{figure}

The study relies on a fairly standard architecture for NLI with two LSTM encoders (contextualization layer) linked through a cross-attention model and a decision layer as illustrated in Fig.~\ref{fig:model}. We make the assumption that the premise and hypothesis sentences have the same statistical attributes: they are thus processed with the same embedding and contextualization layers (i.e., shared weights). The sentence embedding resulting from the contextualization layer could have been used for classification, but we add an attention layer not only to (a) improve the model decision in the classifying layer, but also to (b) benefit from the fact that the attention weight in this position can tell us which elements are taken into consideration by the classifier. 

\begin{figure}[b!]
\centerline{\includegraphics[width=\columnwidth]{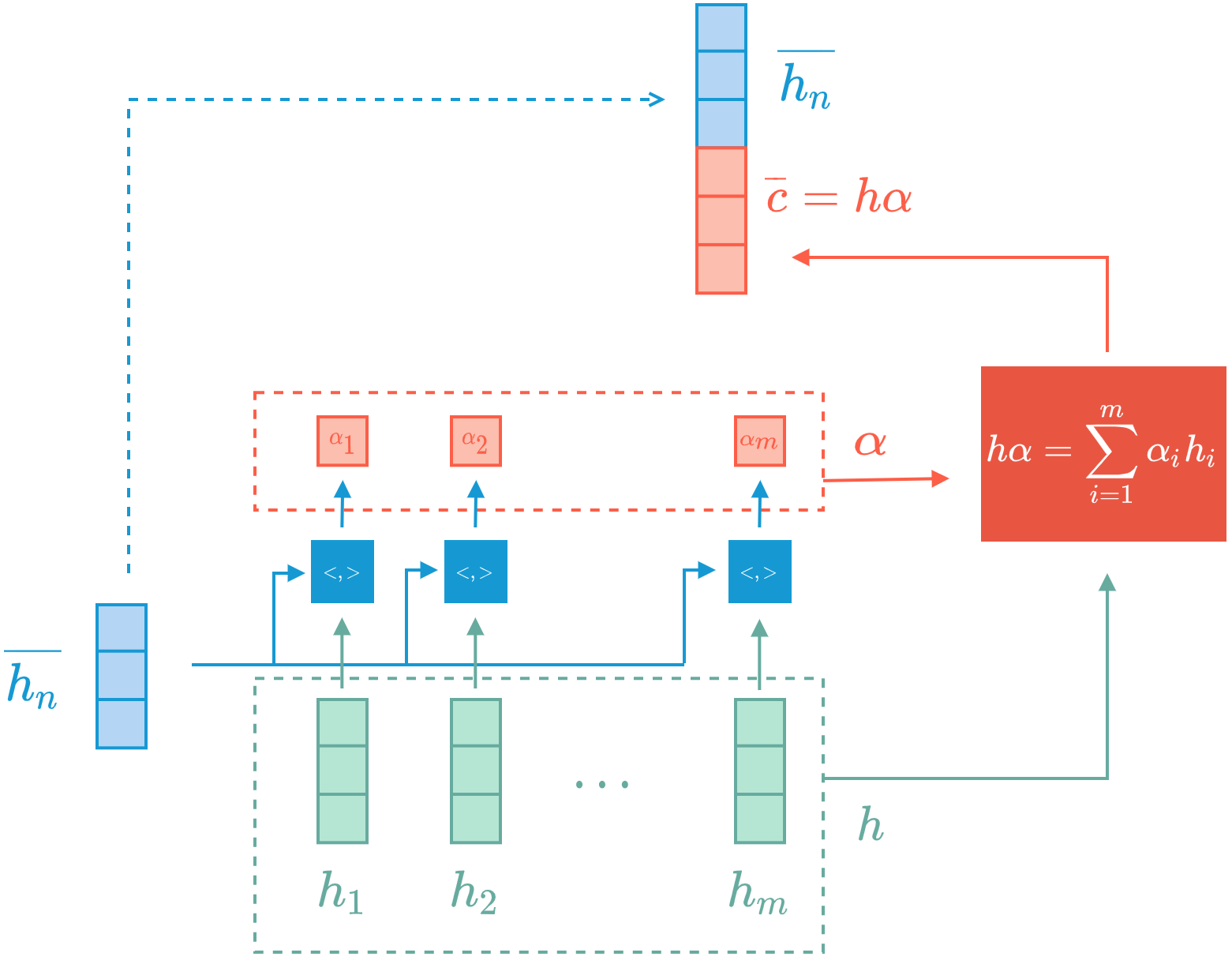}}
\caption{The attention mechanism used in the experimentation. The intuition is that $\alpha$ vector should give how much relevant a word is comparing to the sentence embedding of the other side $\overline{h_m}$.}
\label{fig:attention}
\end{figure}

More precisely, each word in the sentences, after being transformed by a tokenizer, is represented by an embedding vector $w_i$ (and $v_i$ for word vector) through the first layer, $i$ denoting the token position. The embedding layer weight is initialized from pretrained word vectors issued by the GloVe algorithm but we keep the weight trainable. 
A word embedding vector might encode ambiguous information if the context is disregarded. The representation is therefore refined by the context (i.e., the surrounding words) with a LSTM to contextualize word vectors into hidden vectors, denoted $h = [h_1, h_2, ... ,  h_{m}]$ for a sentence of $m$ word tokens, according to
\begin{equation} \label{eq:lstm}
	h_t = \mbox{LSTM}(w_t, h_{t-1}) \enspace .
\end{equation}
We chose to initialize the hidden vectors as null vectors $h_0 = \vec{0}$. 
The last hidden vector $h_m$ is known to capture the sentence meaning and is often referred to as the sentence embedding. If we denote $\overline{h}_n$ the hidden sequence for the remaining text sequence, we can define an attention mechanism that compares each word $h_i$ of one sentence (the premise or the hypothesis) with the sentence embedding $\overline{h}_n$ of the opposite sentence. As in the heuristic, the $\mbox{\it cosine}$ similarity is used to estimate the proximity between $h_i$ and $\overline{h}_n$. 
%
Given the score for each word, the formal attention weight vector $\alpha = [\alpha_1, ..., \alpha_m]$ is a softmax normalized vector to yield a relative magnitude, the definition of the attention weight at each word position $i$ being defined as
\begin{equation} \label{eq:attn}
	\alpha_i = \frac{\displaystyle \exp(h^{\intercal}_i \; \overline{h_n}) }{\displaystyle \sum_{ h_k \in h } \exp(h^{\intercal}_k \; \overline{h_n}) } 
	\enspace .
\end{equation}
In addition, the attention mechanism produces a new enhanced sentence embedding $c$, also referred to as the context vector in~\cite{luong_effective_2015}. This embedding is then concatenated and blended with the RNN embedding $h_n$ by a fully connected layer. The two new merged embeddings are then concatenated for classification. The attention process is illustrated in Fig.~\ref{fig:attention}.

For the classifier part, several fully connected layers are stacked, with ReLU and tanh experimented as activation.  A final softmax is applied at the final layer to classify into 3 classes, with cross-entropy as the loss function in training.

This model is inspired by the decoder-based architecture from~\cite{luong_effective_2015}. We use the last final layer from the hidden representation of a BiLSTM to perform attention against each word in the opposite sequence. As shown in Fig.~\ref{fig:attention}, the dot product is chosen as the attention score, as we need to compute the semantically similar words between the text sequences regarding the task. \autoref{tab:model-params} reports the optimal model hyperparameters by maximizing the F-1 score on the dev set of eSNLI.

\begin{table}[b!]
\begin{center}
\caption{Optimal model configuration.}
\label{tab:model-params}
\begin{tabular}{|c|c|}
\hline
Configuration & optimal setting \\
\hline
Embedding & pretrained GloVe, 300d \\
LSTM & BiLSTM, 1 layer, 300d \\
MLP Classifier & 1 hidden layer, $relu$ activation \\
\hline
\end{tabular}
\end{center}
\end{table}

\section{Experimental Evaluation}
\label{sec:experimentation}

\subsection{Experiment Setup}
\label{exp-setup}

The model is trained on the entire train set of eSNLI with hyperparameters optimized on the dev set. Attention maps are evaluated on the samples from the entailment class from the test set as the idea is to provide explanability on the positive class, i.e., when something of interest is detected.

\subsection{Results}
\label{sec:results}

The first question that we address is whether the model-based map matches the heuristic map or the human-annotated map. Let us first quantitatively analyze the proximity against human annotation.

\begin{figure}[bt!]
\centerline{\includegraphics[width=\columnwidth]{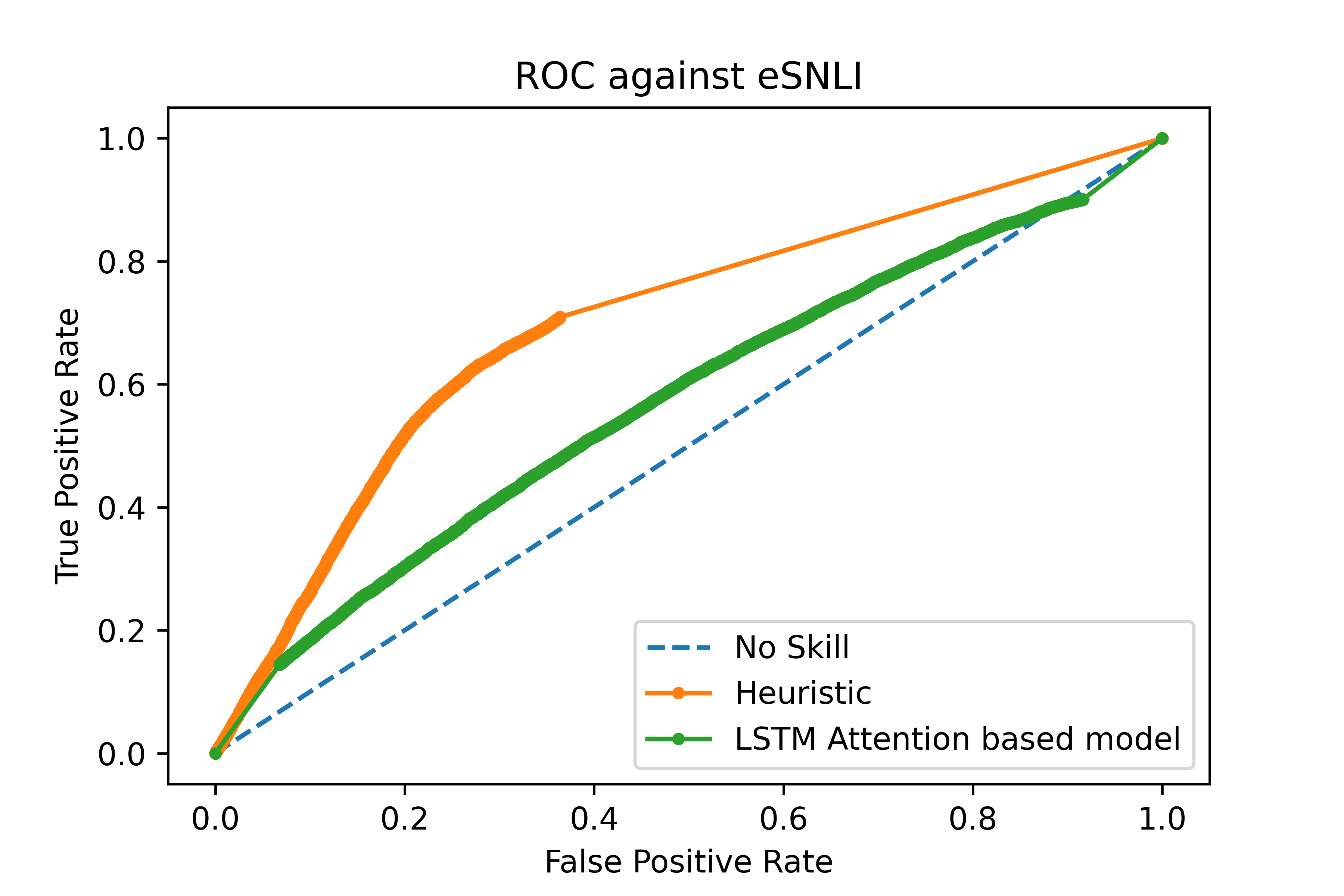}}
\caption{ROC curve that measures how much a model matches the ground truth. The orange (resp. green) line compares the heuristic (resp.\ model-based )map with the human annotation. The dash blue line in diagonal is the worst prediction, where a system highlights every words. }
\label{fig:result_ROC_esnli}
\end{figure}

To compare against the binary human annotation, the attention values in the attention maps, whether model-based or heuristic-based, must be turned into binary values by selecting a threshold $\epsilon \in [0,1]$, the attention values being $1$ above $\epsilon$ and $0$ otherwise. By investigating the true positive and false positive rates as a function of $\epsilon$, we obtain a ROC curve as shown in Fig.~\ref{fig:result_ROC_esnli}. A good matching pair has a high true positive rate in almost all $\epsilon$ values so the corresponding ROC curve leans towards the upper left corner of the graphic. Results in Fig.~\ref{fig:result_ROC_esnli} show a good match between the heuristic map and human annotations. 

Can we conclude the model-based map is closer to the heuristic one? Let us establish a similar curve,  this time comparing the model-based and human-annotated maps against the heuristic map---or, similarly, we use the heuristic map as ground truth. 
However, turning a heuristic map into a ground truth label needs to define a $\epsilon_{\mbox{\tiny heuristic}} \in [0,1]$ as previously. The problem is that for each $\epsilon_{\mbox{\tiny heuristic}}$, there is a different shape of the ROC curves pairs. We thus use the area under the curve (AUC) to measure how relatively high a curve is---the higher the curve, the better the map matches its ground truth---, reporting AUC for different values of $\epsilon_{\mbox{\tiny heuristic}}$.  

\begin{figure}[bt!]
\centerline{\includegraphics[width=0.9\columnwidth]{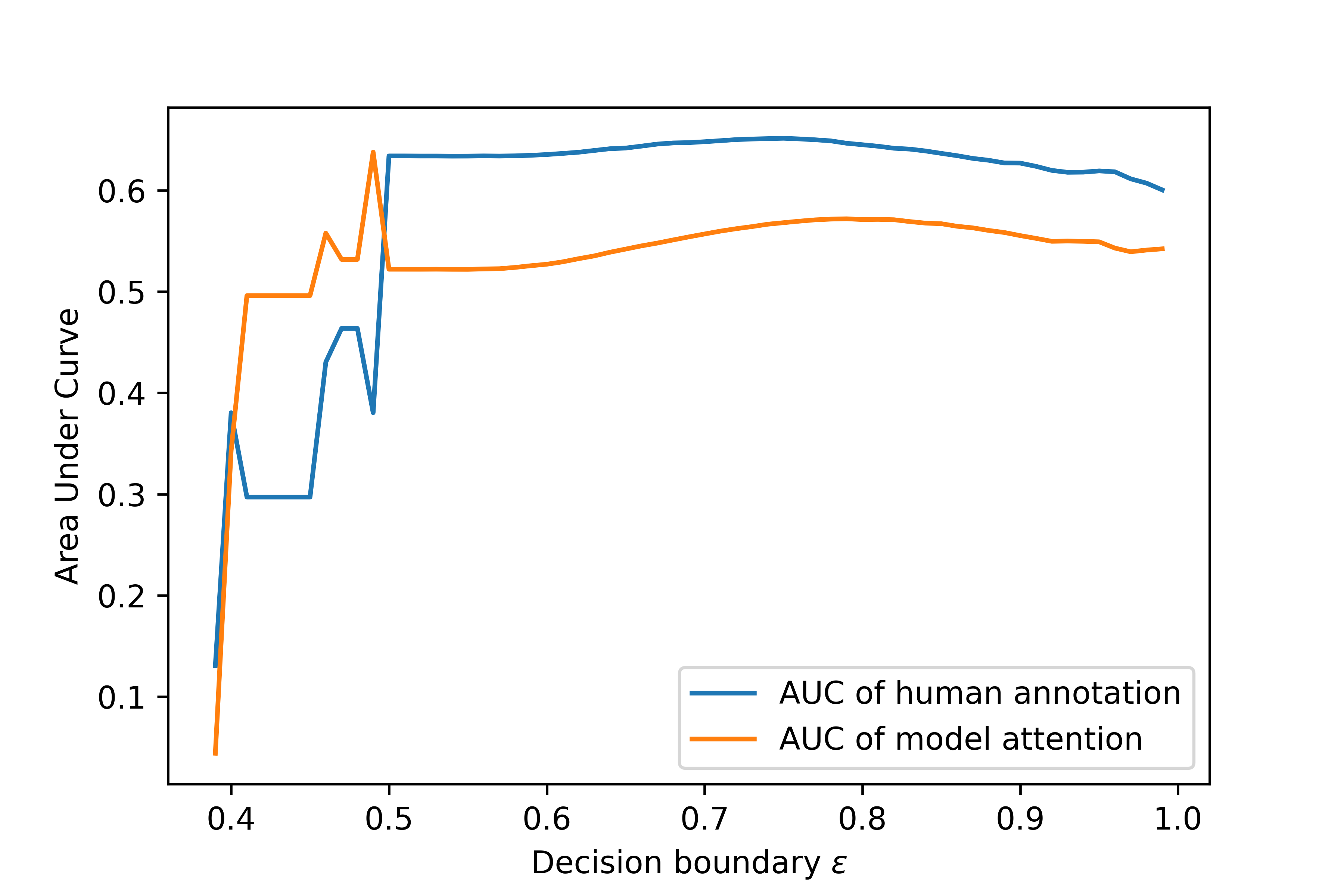}}
\caption{AUC of human annotation map (orange) and model attention (blue) as a function of $\epsilon_{\mbox{\tiny heuristic}}$.}
\label{fig:AUC_across_epsilon}
\end{figure}

\autoref{table:auc} reports the AUC values for some values of $\epsilon_{\mbox{\tiny heuristic}}$. We observe that with $\epsilon_{\mbox{\tiny heuristic}} \in [0.42, 0.5]$, the model attention matches better the heuristic, while overall, the heuristic matches better the annotation. One reason is that at low $\epsilon_{\mbox{\tiny heuristic}}$ values, the heuristic has much more highlights across sentences that cover even parts non highlighted by humans. Meanwhile, model attention tends to distribute highlights on many unimportant tokens. Figure~\ref{fig:AUC_across_epsilon} also shows that for small values of $\epsilon_{\mbox{\tiny heuristic}}$, the model attention matches better the heuristic than the human highlight map. For $85.24\,\%$ of $\epsilon_{\mbox{\tiny heuristic}}$ values, the human annotation matches the heuristic map better than the model attention. We can now answer the first question: the attention map has not enough plausibility compared to neither human nor heuristic ground truth. 

\begin{table}[tb]
\caption{AUC as a function of $\epsilon_{\mbox{\tiny heuristic}}$.}
\label{table:auc}
\begin{center}
\begin{tabular}{|c|c|c|}
\hline
$\epsilon_{heuristic}$ & $AUC_{human}$ & $AUC_{model}$ \\
\hline
0.4 & 0.380450 & 0.342720\\
0.5 & 0.634234 & 0.522256 \\
0.6 & 0.635616 & 0.527185 \\
0.7 & 0.648283 & 0.557091 \\
0.8 & 0.645316 & 0.571382 \\
0.9 & 0.627151 & 0.555492 \\
\hline
\end{tabular}
\end{center}
\end{table}

The next interesting question is whether the heuristic map reflects how a human picks up words to explain the label decision. In other words, does the distribution of attention values (actual values, not thresholded binary values) match between the heuristic and the human annotations. For the latter, binary values are normalized so that attention values sum to unity in each sentence. On a quantitative global scale, we measure the correlation between the distribution of attention values between the two maps in \autoref{table:global_correlation}. The small values of the p-value for both Pearson and Spearman correlation reject the null hypothesis $H_0$ that the correlation is statistically insignificant, i.e., the correlation values show a low correlation between the two maps. 

\begin{table}[tb]
\begin{center}

\caption{Correlation between heuristic and human annotation.}
\label{table:global_correlation}
\begin{tabular}{|c|c|c|c|c|}
\hline
 & \multicolumn{2}{|c|}{\textbf{Pearson}} & \multicolumn{2}{|c|}{\textbf{Spearman} } \\
\cline{2-5} 
 & Correlation & p-value & Correlation & p-value \\
\hline
Premise & 0.5227 & $< 10^{-300}$ & 0.5337 & $< 10^{-300}$ \\
Hypothesis & 0.5227 & $< 10^{-300} $ & 0.4605 & $< 10^{-300}$ \\
\hline
General & 0.5206 & $< 10^{-300}$ & 0.4426 & $< 10^{-300}$ \\
\hline
\end{tabular}
\end{center}
\end{table}

\begin{figure}[b!]
\centerline{\includegraphics[width=\columnwidth]{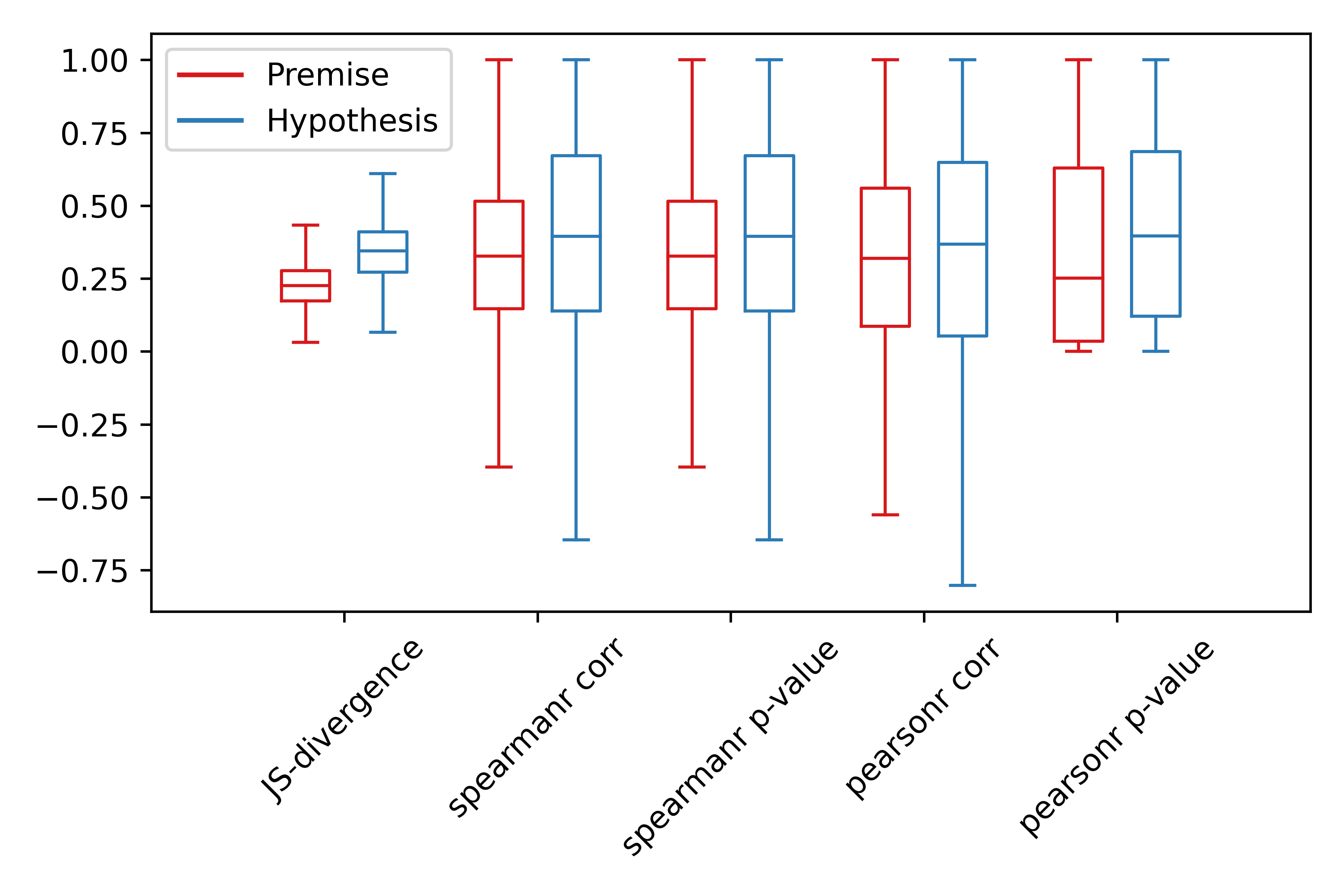}}
\caption{Distribution of Spearman and Pearson correlation ratios across sentences (premise in red, hypothesis in blue).}
\label{fig:correlation_per_instance}
\end{figure}

However, the heuristic map at a per instance scale still has a dispersion across sentence pairs as shown in Fig.~\ref{fig:correlation_per_instance} where we report the distribution of several metrics measured at the sentence level rather than at the global level. We report three different metrics: the Jensen-Shannon divergence (JS-divergence) between the two distributions of attention values (the lower, the better), as well as the Spearman and Pearson correlations (the higher, the better). The distribution of the JS-divergence in Fig.~\ref{fig:correlation_per_instance} shows that most of the distribution significantly differ. Similarly, the high p-values in Spearman and Pearson correlation tests reject any correlation between the two distributions.

\begin{figure*}[tb!]
\centerline

\subfloat[Short pair]{\includegraphics[width=0.9\columnwidth]{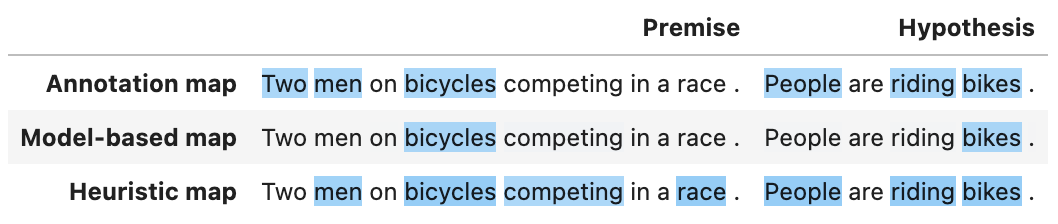}}
\subfloat[Long pair]{\includegraphics[width=1.1\columnwidth]{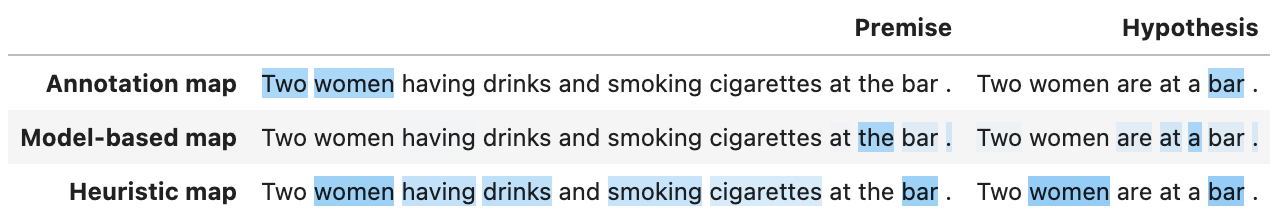}}
\caption{Two examples comparing human-based, model-based and heuristic-based attention maps.}
\label{fig:qualtitative}
\end{figure*}

A qualitative investigation can confirm this last observation. We show examples of two sequence pairs of the entailment class in Fig.~\ref{fig:qualtitative}, where the heuristic can provide better coverage on human annotation highlights, while the model-based one mismatches many words. This result explains why, on a global scale, the heuristic still has a relation with human annotation, though not a perfect match.

In summary, the attention between RNN encoders has a very low plausibility, even compared to the heuristic map; one reason is that the model focuses mostly on unimportant grammatical parts of the sentence (determinants, punctuation, etc.) and skips the informative syntactic groups for human (verb, noun, adjective). On the other hand, the heuristic is proved to be a good baseline to measure plausibility in NLI task. Confirming~\cite{tutek_staying_2020}, the embedding presentation can provide useful information for word choice and may result in higher plausibility in the heuristic map in NLI task.

\section{Conclusion}
\label{sec:conclusion}

In this paper, we performed an empirical study on the plausibility of the attention mechanism in a sentence comparison task. The attention map is built on top of RNN-encoders---BiLSTM models. In this work, we exploited human explaining annotations from the eSNLI corpus. The explanations in the corpus are provided by an explaining phrase for the label and highlighted maps. We used the latter as a ground truth to evaluate the model attention map. In addition, we designed a heuristic method, based on words similarity and grammatical attributes, as an additional second ground-truth to assess the plausibility.
The results showed that the model-based attention map is not close to both ground-truth maps. The qualitative analysis shows that the model gives scattered attention weights, most of them on stop words, which bring few indicative information to humans. However, the heuristic map, based on initial embedding words, shows a better coverage of these words. 

This observation shows the importance of the word embedding vectors which could bring useful information to the attention mechanism to improve its plausibility. In future works, the heuristic could be used as a tool for evaluation in NLI tasks for corpus without annotations. As an extension, the heuristic could serve as a complementary to a regularization proposed in~\cite{tutek_staying_2020} to enhance the plausibility without sacrificing the faithfulness of the attention mechanism.


\bibliographystyle{IEEEtran}
\bibliography{./my-paper}

\end{document}